\documentclass{article}

\usepackage{arxiv}

\usepackage[utf8]{inputenc} 
\usepackage[T1]{fontenc}    
\usepackage{hyperref}       
\usepackage{url}            
\usepackage{booktabs}       
\usepackage{amsfonts}       
\usepackage{nicefrac}       
\usepackage{microtype}      
\usepackage{lipsum}		
\usepackage{graphicx}
\usepackage{natbib}
\usepackage{doi}

\usepackage{algorithmic}
\usepackage{algorithm}
\usepackage{multirow}
\usepackage{subcaption}

\title{ProtoCAM: Interpretable Few-Shot Mask-Guided Prototypical Learning for Breast Lesion Classification in Ultrasound Imaging}

\date{} 					

\author{ \href{https://orcid.org/0000-0002-4542-9105}{\includegraphics[scale=0.06]{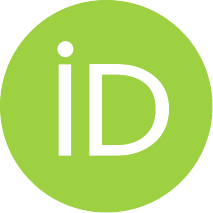}\hspace{1mm}Ashkan~Ebadi}\\
	Digital Technologies Research Centre\\
	National Research Council Canada\\
	Toronto, ON M5T 3J1 \\
	\texttt{ashkan.ebadi@nrc-cnrc.gc.ca} \\
}

\renewcommand{\shorttitle}{\textit{arXiv} Template}

\hypersetup{
pdftitle={ProtoCAM},
pdfauthor={Ashkan~Ebadi},
}

\begin{document}
\maketitle

\begin{abstract}
Breast ultrasound imaging plays an important role in the early detection and diagnosis of breast cancer, particularly for patients with dense breast tissue. However, developing reliable deep learning models for ultrasound analysis is challenging due to limited annotated medical data and the need for interpretable predictions. To address these challenges, this paper proposes ProtoCAM, an explainable few-shot learning framework for breast lesion classification that integrates mask-guided feature encoding, prototypical metric learning, and gradient-based visual explanations. The proposed approach leverages lesion masks to guide feature extraction and constructs class prototypes within an embedding space to enable robust classification under limited training samples. The framework was evaluated on the BUSI dataset using a stratified group k-fold cross-validation protocol to prevent patient-level data leakage. Experimental results demonstrate ProtoCAM's high performance in low-data scenarios. In a 3-way 5-shot setting, the proposed method achieves a macro F1-score of 0.910, representing a substantial improvement over standard supervised CNN models. Among the evaluated backbone networks, ResNet18 achieved the best performance, reaching a macro F1-score of 91.65 percent under a 15-shot configuration, providing interpretable insights into the classification decisions. These results highlight the potential of explainable few-shot learning frameworks for reliable computer-aided breast cancer diagnosis in data-scarce medical imaging environments.
\end{abstract}


\section{Introduction}
Cancer is a pathological condition in which cells acquire the ability to proliferate uncontrollably, bypassing the normal regulatory mechanisms that govern cell division and apoptosis \cite{national_cancer_institute_what_2007}. In breast cancer, this abnormal growth originates in the epithelial cells of the breast ducts or lobules, leading to the formation of malignant tumours. These cancerous cells can invade surrounding breast tissue and, in advanced stages, spread to other parts of the body via the lymphatic system or bloodstream \cite{world_health_organization_breast_2025,singh_breast_2024}. Breast cancer is the most frequently diagnosed cancer among women globally, with over 2.3 million cases reported in 2022 \cite{world_health_organization_breast_2025}. It remains one of the leading causes of cancer-related mortality in women, accounting for approximately 670,000 deaths, around 7\% of all cancer deaths worldwide \cite{world_health_organization_breast_2025}. Incidence rates show marked geographic variation, exceeding 80 cases per 100,000 women in developed regions such as Europe, North America, Australia, and New Zealand \cite{arnold_current_2022}.

Early and accurate detection of breast cancer is critical for improving survival outcomes \cite{seely_screening_2018}, as it enables the timely initiation of effective treatment. While conventional diagnostic modalities—such as mammography, ultrasound, and biopsy—are widely used, each presents inherent limitations, including operator dependence, inter-observer variability, and the need for specialized expertise. Biopsy, although regarded as the diagnostic gold standard, is invasive and can cause patient discomfort \cite{upadhyay_enhancing_2025}. Ultrasound (US) imaging is a non-invasive, well-tolerated, real-time diagnostic technique commonly used in the evaluation of breast abnormalities \cite{zhang_shamtl_2021}. Owing to its accessibility, absence of ionizing radiation, and ability to differentiate between cystic and solid lesions, it is often preferred as an initial assessment tool in breast cancer diagnosis over modalities such as mammography or biopsy \cite{isik_fewshot_2024}.

Computer-aided diagnosis (CADx) systems play a vital role in the early detection of breast cancer and other malignancies \cite{isik_fewshot_2024}. Traditional CAD solutions based on classical machine learning often struggled to generalize across diverse populations and imaging conditions, whereas deep learning–based approaches have demonstrated substantial improvements by automatically learning complex, discriminative features from raw data. Recent studies applying deep learning to breast cancer detection in ultrasound images have reported promising results \cite{sharafaddini_deep_2025}; however, their performance is frequently constrained by the limited availability of large, diverse medical imaging datasets, which are essential for building robust and generalizable models \cite{archana_deep_2024}. Medical imaging datasets are typically scarce due to several factors, such as privacy restrictions, high acquisition costs, and the difficulty of obtaining expert-labelled samples \cite{upadhyay_enhancing_2025,ebadi_2026}. In such data-scarce scenarios, few-shot learning (FSL) offers a compelling alternative, enabling models to adapt to new cases and classes using only a small number of labelled examples \cite{snell_prototypical_2017}. Metric learning strategies are particularly well-suited for this task, as they learn embedding spaces that capture meaningful similarities between samples \cite{song_covidnet_2023}, facilitating accurate classification even in low-resource settings.

Another critical challenge is the inherent opacity of deep learning models, often described as ``black boxes'', whose internal decision-making processes are difficult to interpret even for domain experts \cite{van_der_velden_explainable_2022}. In medicine, explainability is not merely a technical preference but a clinical necessity, essential for fostering trust among practitioners, meeting regulatory requirements, and ensuring ethical accountability \cite{houssein_explainable_2025}. Explainable artificial intelligence (XAI) offers a means to address these challenges by producing interpretable outputs that link artificial intelligence (AI) predictions to clinical reasoning. Techniques such as visual heatmaps in medical images or textual justifications help highlight the features influencing a model’s decision, thereby enhancing transparency and trust in AI-assisted diagnosis \cite{houssein_explainable_2025}.

This work presents a high-performing mask-guided few-shot solution for ultrasound lesion classification, called ProtoCAM, by integrating data-driven metric learning with rigorous validation and explainability assessments. By employing prototypical networks \cite{snell_prototypical_2017} optimized with dataset-specific normalization, the pipeline overcomes the severe data scarcity and low-contrast challenges inherent to few-shot medical imaging, systematically outperforming standard convolutional neural network (CNN) baselines. The solution seamlessly embeds Grad-CAM \cite{selvaraju2017grad} feature visualizations to verify that the model's performance is driven by genuine pathological regions, such as irregular tumour margins, rather than background artifacts, providing an auditable end-to-end architecture tailored for reliable medical assessment and deployment.

The main contributions of this work are summarized as follows. First, we propose ProtoCAM, an explainable few-shot learning framework for breast ultrasound image classification that integrates mask-guided feature encoding with prototypical metric learning to address the challenge of limited annotated medical data. Second, we incorporate lesion-aware feature extraction using ground-truth masks provided in the dataset to guide the network toward clinically relevant regions, improving representation quality and classification performance. Third, we conduct a comprehensive experimental evaluation on the BUSI dataset using a stratified group K-fold cross-validation protocol to prevent patient-level data leakage and ensure reliable performance assessment. Finally, we provide model interpretability through Grad-CAM visualizations, demonstrating that the proposed framework focuses on meaningful lesion regions and produces clinically interpretable explanations for its predictions. 

The remainder of this paper is organized as follows. Section~\ref{sec:data} describes the dataset used in this study; Section~\ref{sec:method} describes the proposed ProtoCAM framework; Section~\ref{sec:results} presents the experimental setup and reports and discusses the results; Section~\ref{sec:conclusion} concludes the paper; and Section~\ref{sec:limit} highlight the limitations and outlines future research directions.

\section{Data}\label{sec:data}
In this study, we used the Breast Ultrasound Images (BUSI) dataset \cite{BUSI2020dataset}, which comprises breast ultrasound images collected in 2018 from 600 female patients aged 25 to 75 years. The collection includes a total of 780 images, each with an average resolution of $500\times500$ pixels and stored in PNG format. For each original image, corresponding ground truth annotations (for some cases, more than one) are provided to support evaluation. The dataset is organized into three diagnostic categories: normal, benign, and malignant. Table~\ref{tab1:data_dist} shows the distribution of images. These images capture a wide range of breast tissue characteristics, making the dataset suitable for advanced analytics.

\begin{table}
\caption{Distribution of images across diagnostic categories}
\begin{center}
\begin{tabular}{r@{\quad}l@{\quad}l} 
\hline
Category & Description& \#Images \\
\hline
Normal& Ultrasound images of healthy breast tissue& 133 \\
Benign& Images showing non‑cancerous breast lesions& 437 \\
Malignant& Images depicting cancerous breast lesions& 210 \\
\hline
Total&  & 780 \\
\hline
\end{tabular}
\label{tab1:data_dist}
\end{center}
\end{table}

\section{Methodology}\label{sec:method}
This study proposes a few-shot learning framework for breast cancer classification from ultrasound images using a prototypical network architecture. The framework integrates lesion-aware feature extraction, episodic meta-learning, and patient-wise cross-validation to address the limited availability of labelled medical data. The methodology is described in detail in this section.

\subsection{Data Preprocessing}
All images undergo resizing and normalization prior to model training. Images are resized to a fixed square resolution, with experiments conducted using sizes ranging from 224 to 320 pixels.\footnote{Due to space limitations, we report results only for an image size of $288 \times 288$.} Pixel intensities are normalized using dataset-specific statistics. Masks are resized using nearest-neighbour interpolation and binarized to preserve lesion boundaries.

\subsection{Episodic Learning Framework}
The learning paradigm follows an episodic training strategy. Instead of conventional mini-batches, the model is trained on episodic tasks designed to mimic the few-shot inference scenario. Each episode is constructed as an N-way K-shot classification task. In this work, the primary configuration uses a three-way classification problem corresponding to the three diagnostic classes. For each episode:

\begin{itemize}
    \item K-shot support samples are drawn for each class.
    \item Q query samples are drawn for evaluation within the episode.
\end{itemize}

The default experimental configuration uses 5-shot support samples and 20 query samples per class, though experiments were conducted with larger support sizes such as 10-shot, 15-shot, and 20-shot settings. Episodes are generated using a custom episodic sampler that ensures balanced sampling across classes.

\subsection{Data Augmentation}
To improve generalization and mitigate overfitting in the few-shot setting, a set of paired data augmentations is applied to both the image and the corresponding lesion mask during training as part of the episodic learning process. These augmentations include: 1) Horizontal flipping with probability 0.5, 2) Vertical flipping with probability 0.2, 3) Random rotation within $\pm$25 degrees, 4) Random affine transformation including translation up to 10\% of image size and scaling between 0.9 and 1.1, and 5) Colour jitter applied to image brightness and contrast. These transformations are applied identically to both the image and its mask to maintain spatial alignment. Validation data undergo deterministic preprocessing without random augmentation.

\subsection{Architecture}
The core model follows the prototypical network architecture, which learns a metric space in which classification is performed by computing distances to class prototypes. 

\subsubsection{Backbone Feature Extractor}
Feature representations are extracted using pre-trained convolutional neural network backbones. The classification layers of these networks are removed, and only the feature embeddings are used. Several architectures were evaluated, including: ResNet18, ResNet50, ResNet101, MobileNetV2, DenseNet121, EfficientNetB0, EfficientNetB1, and VGG16. Among these, ResNet18 served as the primary backbone due to its favourable balance between representation capacity and computational efficiency.

\subsubsection{Mask-Guided Feature Encoding}
A key component of the framework is lesion-focused embedding generation using spatial mask gating, producing lesion-centric embeddings that emphasize diagnostically relevant tissue structures. This approach improves representation quality by reducing background noise in ultrasound images.

\subsubsection{Prototype-Based Classification}
\begin{figure}[htbp]
\centering
\includegraphics[scale=0.52]{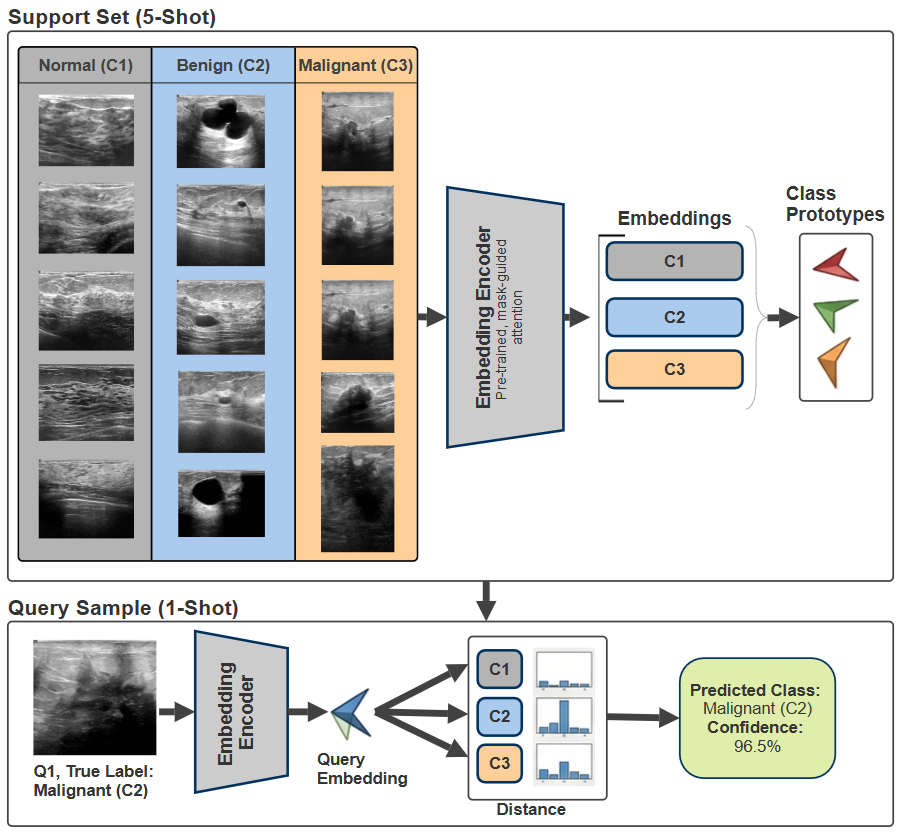}
\caption{Overview of the mask-guided few-shot learning pipeline for breast ultrasound classification under a five-shot scenario.}
\label{fig1}
\end{figure}

For each episode, class prototypes are computed by averaging the embeddings of the support samples belonging to the same class. If $z_i$ represents the embedding of support sample $i$ from class $c$, the prototype for class $c$ is computed as the mean embedding of its support examples. Query embeddings are then compared to these prototypes using a distance metric. We experimented with both Euclidean distance and cosine similarity, and Euclidean distance performed better. Figure~\ref{fig1} shows the high-level architecture for a three-way five-shot scenario. 

\subsection{Training Strategy}
The model is trained using episodic batches across multiple epochs. During each training episode: 1) Support and query images are passed through the backbone encoder to produce embeddings, 2) Class prototypes are computed from support embeddings, 3) Query embeddings are compared against prototypes, and 4) The distances are converted into class probabilities. Model optimization is performed using the Adam optimizer with an initial learning rate of $5 \mathrm{e}{-5}$ and an initial weight decay of $1 \mathrm{e}{-3}$. The models are trained and tested over 10 to 80 epochs, depending on the backbone, setting, and parameter ranges. A \texttt{ReduceLROnPlateau} scheduler monitors validation loss and reduces the learning rate by a factor of 0.1 if the validation loss does not improve for 4 consecutive epochs.

\subsection{Cross-Validation Strategy and Evaluation Metrics}
To ensure robust evaluation and prevent patient-level data leakage, a Stratified Group K-Fold cross-validation scheme is employed. This strategy preserves class distribution across folds while ensuring that all images from the same patient remain within a single fold. The number of splits is set to three folds. For each fold, the dataset is partitioned into training and validation subsets and episodic sampling is performed independently within each subset. The best-performing model is selected based on validation Macro F1-score, which balances both false positives and false negatives independently for each class and then averages them equally. This prevents the model from hiding behind overall accuracy by simply predicting the majority class, which in breast cancer screening would be a potentially life-threatening blind spot. Model checkpoints are stored whenever a new best validation performance is observed. The final trained model is also serialized for future analysis.

\subsection{Episodic Few-Shot Learning Formulation}
The learning process is formulated as episodic training where each episode simulates an N-way K-shot classification task. Let the dataset be defined as: $D = {(x_i, y_i)}$ for $i = 1$ to $M$, where $x_i$ represents an ultrasound image and $y_i \in {1,\dots, C}$ denotes the class label. During training, an episode $E$ is constructed by randomly sampling $N$ classes and selecting $K$ support samples and $Q$ query samples for each class. The support and query set ($S$ and $Q$, respectively) are defined as: $S = {(x_i, y_i)}$ for $i = 1$ to $N \times K$, and $Q = {(x_j, y_j)}$ for $j = 1$ to $N \times Q$. The support set is used to compute class prototypes, while the query set is used to evaluate the model within the episode.

Each ultrasound image is passed through a convolutional backbone encoder $f\theta(\cdot)$ (e.g., ResNet18) to produce feature maps: $F = f\theta(x)$. Instead of global average pooling over the entire feature map, our proposed framework uses lesion-aware masked pooling guided by the mask $m$. The mask is resized to match the spatial dimensions of the feature map. Let $F(h, w, c)$ denote the activation at spatial location $(h, w)$ for channel $c$, and $m(h, w)$ denote the corresponding binary lesion mask. The masked embedding vector $z$ is computed as:

\begin{equation}
z = \frac{\Sigma_h \Sigma_w F(h,w,:) \times m(h,w)}{\Sigma_h \Sigma_w m(h,w)}
\end{equation}

This operation ensures that the resulting representation emphasizes lesion regions while suppressing irrelevant background tissue. To stabilize the embedding space, the resulting vector is normalized using $L2$ normalization. For each class $c$ in the support set, a prototype vector is computed as the mean of the embeddings of its $K$ support samples. Given a query embedding, classification is performed by computing its distance to each class prototype. Euclidean distance is used as the distance measure. The classification logits are computed as the negative distance scaled by a learnable temperature parameter $\tau$ that controls the sharpness of the probability distribution. The probability that the query sample belongs to class $c$ is obtained using the softmax function.

Model parameters $\theta$ are optimized by minimizing the negative log-likelihood loss computed over the query samples of each episode. For a query sample $(x_q,y_q)$, the loss is defined as:

\begin{equation}
    \mathcal{L} = -\log P(y_q | x_q)
\end{equation}

For an episode containing $N \times Q$ query samples, the total loss becomes:

\begin{equation}
    \mathcal{L}_{episode} = \frac{1}{N \times Q} \times \Sigma_{q} -\log P(y_q|x_q)
\end{equation}

This episodic objective encourages the encoder to learn a feature space in which samples from the same class cluster around their corresponding prototypes while remaining separated from other classes. 

The model parameters are optimized using the Adam optimizer with learning rate $\alpha$ and weight decay $\lambda$. During training, each episode proceeds as follows: 1) Sample $N$ classes and construct support and query sets, 2) Compute support embeddings using the lesion-aware encoder, 3) Compute class prototypes by averaging support embeddings, 4) Encode query images and compute distances to prototypes, 5) Compute softmax probabilities and episodic loss, and 6) Update model parameters via backpropagation. This episodic training strategy enables the network to learn a generalizable metric space suitable for few-shot classification tasks. The high-level episodic training procedure is outlined in Algorithm 1.

\begin{algorithm}[htbp] 
\caption{Episodic Training for ProtoCAM} 
\begin{algorithmic}[1] 
 \renewcommand{\algorithmicrequire}{\textbf{Input:}}
 \renewcommand{\algorithmicensure}{\textbf{Output:}}
\REQUIRE Dataset $D={(x_i,y_i,m_i)}$, \\
         encoder $f_{\theta}$, \\
         $N$-way $K$-shot, \\
         $Q$ queries per class, \\
         episodes $E$ 
\ENSURE Trained parameters $\theta$

\FOR{$e = 1$ to $E$} 
\STATE Sample $N$ classes from $D$ 
\STATE For each class $c$, sample $K$ support and $Q$ query images 
\STATE Construct support set $S$ and query set $Q$
\STATE Compute support embeddings $z_s = f_{\theta}(x_s, m_s)$ 
\STATE Compute query embeddings $z_q = f_{\theta}(x_q, m_q)$
    \FOR{each class $c$} 
        \STATE Compute prototype: $p_c = \frac{1}{K}\sum z_s^c$ 
    \ENDFOR
    
    \FOR{each query embedding $z_q$} 
    \STATE Compute distances $d_c = ||z_q - p_c||_2$ 
    \STATE Compute class probabilities using softmax over $-d_c$ 
    \ENDFOR
\STATE Compute episodic loss $\mathcal{L}$ on query labels 
\STATE Update $\theta$ using Adam optimizer 
\ENDFOR 
\end{algorithmic} 
\end{algorithm}

\subsection{Model Interpretability}
Grad-CAM \cite{selvaraju2017grad} visualizations are generated to improve interpretability. Grad-CAM is applied to the final convolutional layer of the backbone network. Heatmaps are overlaid on the original ultrasound images to visually assess whether the model focuses on relevant lesion regions.

\subsection{Computational Environment and Implementation Details}
All experiments were implemented in Python 3.14 using the PyTorch deep learning framework. Training was conducted on a workstation equipped with an NVIDIA RTX A6000 GPU and 128 GB RAM. Random seeds were fixed to ensure reproducibility across experimental runs. Model checkpoints, training logs, and evaluation metrics were automatically recorded during training to facilitate experiment tracking and analysis.

\section{Experiments and Results}\label{sec:results}

\subsection{Comparative Analysis of Backbone Models}
To investigate the impact of feature extraction on the proposed few-shot framework, several convolutional neural network backbones were evaluated within the prototypical network architecture under a 15-shot, 25-query configuration. All models were trained and evaluated under the same episodic training configuration to ensure a fair comparison. Table~\ref{tab1:backbones} summarizes the performance obtained using different backbone networks, including ResNet18, ResNet50, ResNet101, DenseNet121, VGG16, MobileNetV2, and EfficientNetB1. Performance is reported in terms of macro-averaged F1-score, recall, precision, and the total number of trainable parameters.

\begin{table}[htbp]
\caption{Performance of alternative backbone models}
\begin{center}
\begin{tabular}{r@{\quad}l@{\quad}l@{\quad}l@{\quad}l@{\quad}}
\hline
Backbone&\multicolumn{4}{c}{Performance Metric} \\
Model & Macro F1 \% & Recall \% & Prec. \% & \#Params\\
\hline
ResNet18& \textbf{91.65}$\pm2.38^{\mathrm{a}}$  & \textbf{91.65} & \textbf{91.66} & 11,176,513  \\
ResNet50& 90.70$\pm1.22$& 90.70 & 90.71 &  23,508,033\\
ResNet101& 89.71$\pm1.38$& 89.71 & 89.71 & 42,500,161 \\
DenseNet121& 90.18$\pm2.28$& 90.19 & 90.19 & 6,953,857 \\
VGG16& 89.50$\pm1.66$& 89.50 & 89.50 & 14,714,689 \\
MobileNetV2& 89.74$\pm1.80$& 89.74 & 89.74 &  2,223,873\\
EfficientNetB1& 90.69$\pm2.30$& 90.70 & 90.70 & 6,513,185 \\
\hline
\multicolumn{5}{l}{$^{\mathrm{a}}$The best-performing value is shown in bold.}
\end{tabular}
\label{tab1:backbones}
\end{center}
\end{table}

Among the evaluated models, ResNet18 achieved the best overall performance, obtaining a macro F1-score of 91.65 percent with a standard deviation of 2.38 across validation folds. It also achieved the highest recall and precision values. Despite being a relatively lightweight architecture compared to deeper variants such as ResNet50 and ResNet101, ResNet18 demonstrated superior generalization in the few-shot learning setting. Deeper networks did not yield improved performance, likely due to the limited number of training samples available in the episodic learning framework. Models with larger parameter counts, such as ResNet101 and VGG16, showed slightly lower performance, which may be attributed to increased risk of overfitting. Conversely, lightweight models such as MobileNetV2 achieved comparable results while using significantly fewer parameters. Another observation is the near-equality of the three metrics, which indicates balanced classification behaviour across the three classes.

\subsection{Sensitivity Analysis of Episodic Parameters}
To examine the influence of episodic sampling parameters on model performance, experiments were conducted with varying support set sizes (K-shot) and query sizes while using the ResNet18 backbone. Specifically, four support configurations (5-shot, 10-shot, 15-shot, and 20-shot) were evaluated with two query sizes (20 and 25 samples per class). The results are summarized in Table~\ref{tab3:shots}. Overall, the results indicate that ProtoCAM achieves consistently strong performance across all configurations, with macro F1-scores ranging from approximately 90.8\% to 91.7\%. Increasing the number of support samples generally improves performance up to a certain point. For example, performance improves from 90.83\% in the 5-shot setting to 91.65\% in the 15-shot setting when the query size is 25. This suggests that providing additional support examples allows the model to construct more representative class prototypes, improving the separability of the learned embedding space.

\begin{table}[htbp]
\caption{Performance of ProtoCAM (ResNet18 backbone) under different K-shot and query configurations}
\begin{center}
\begin{tabular}{r@{\quad}l@{\quad}l@{\quad}l@{\quad}l@{\quad}}
\hline
Number & Query &\multicolumn{3}{c}{Performance Metric} \\
of Shots & Size & Macro F1 \% & Recall \% & Precision \%\\
\hline
\multirow{2}{*}{5}& 20 & 90.97$\pm2.25$  & 90.97 & 90.97 \\
 & 25 & 90.83$\pm1.67$  & 90.83 & 90.83 \\
\multirow{2}{*}{10}& 20 & 90.86$\pm1.74$& 90.86 &  90.86 \\
& 25 & 91.24$\pm2.85$& 91.24 & 91.24 \\
\multirow{2}{*}{15}& 20 & 91.15$\pm2.19$& 91.16  & 91.16 \\
& 25 & \textbf{91.65}$\pm2.38^{\mathrm{a}}$& \textbf{91.65} & \textbf{91.66} \\
\multirow{2}{*}{20}& 20 & 90.82$\pm1.68$& 90.82 & 90.83 \\
& 25 & 91.25$\pm2.08$& 91.25 & 91.25 \\
\hline
\multicolumn{5}{l}{$^{\mathrm{a}}$The best-performing value is shown in bold.}
\end{tabular}
\label{tab3:shots}
\end{center}
\end{table}

The best performance is obtained with the 15-shot, 25-query configuration, which achieves a macro F1-score of 91.65\%. This configuration likely provides a good balance between prototype stability and training diversity for the examined dataset. With 15 support samples per class, the class prototypes are sufficiently representative and increasing the support set size may have introduced near-duplicate examples that do not provide meaningful new information for prototype estimation. In addition, larger support sets can alter the balance between support and query samples within an episode. Since the optimization objective of prototypical networks is computed on the query set, increasing the support size without proportionally increasing query samples may reduce the relative amount of training signal available for gradient updates. The effect of increasing query size from 20 to 25 samples is generally positive but modest. Larger query sets provide more examples for computing the episodic loss, which can lead to more stable gradient estimates during training. This effect is particularly noticeable in the 10-shot and 15-shot configurations, where increasing the query size slightly improves the macro F1-score. 

\subsection{Domain-Specific vs. General-Purpose Pre-training}
To evaluate the impact of domain-specific feature representations on low-contrast medical data, we compared a ResNet50 backbone initialized with weights from HuggingFace RadImageNet \cite{mei2022radimagenet}, a large-scale radiologic imaging dataset, against a ResNet50 baseline initialized with traditional ImageNet natural image weights\footnote{RadImageNet does not offer pretrained weights for ResNet18, therefore, we used ResNet50 in this experiment.}. Empirically, the RadImageNet-initialized prototypical network achieved a Macro F1-score of 87.87\% at 40 epochs, which steadily climbed to a peak of 90.79\% at 80 epochs before succumbing to overfitting at 100 epochs (dropping to 89.93\%). Unexpectedly, the standard ImageNet initialization outperformed the peak medical-specific pre-training by a noticeable margin, reaching a Macro F1-score of 90.96\% in 40 epochs.

Several factors may explain why the medical domain-specific model did not yield superior performance under these few-shot parameters. First, while RadImageNet contains a vast library of multiple medical modalities, these modalities primarily feature high-contrast structural landmarks and sharp bone-to-tissue boundaries. In contrast, ultrasound images are highly unique, dominated by severe speckle noise, low-contrast shadowing, and localized acoustic variations. Second, this may highlight the critical role of ``feature diversity'' versus ``feature specificity'' in few-shot metric spaces. Natural image datasets, e.g., ImageNet, force a network to learn highly versatile, diverse primitives such as curves, textures, scales, and fine gradients across various heterogeneous classes. This expansive, generalized geometric vocabulary may provide a robust, adaptable canvas for the episodic sampler to construct flexible metric boundaries. Conversely, RadImageNet weights are highly specialized toward fixed medical target distributions. When forced into an ultra-low sample domain (3-way 5-shot), this hyper-specificity restricted the prototype space, causing the model to adapt slowly (requiring 80 epochs to reach its peak) and making it highly vulnerable to episodic feature collapse and overfitting shortly thereafter. 

\subsection{Comparative Analysis: ProtoCAM vs. Conventional CNNs}
\begin{figure}[htbp]
\centering
\includegraphics[scale=0.44]{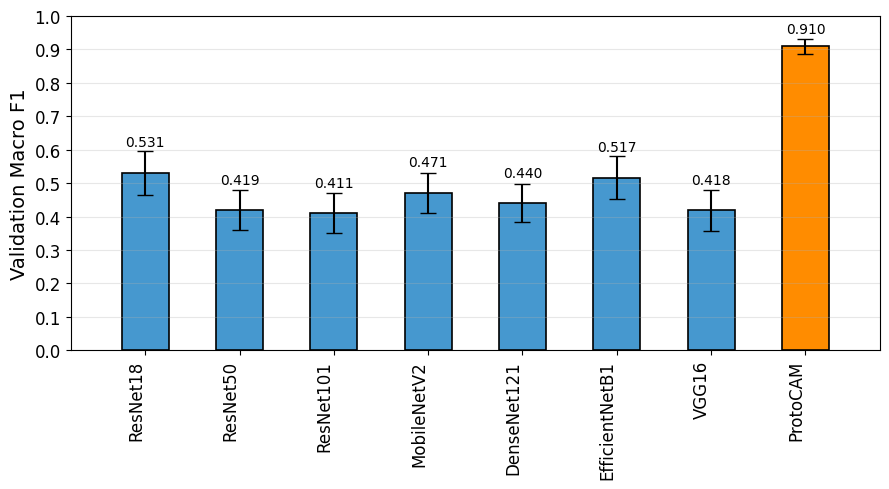}
\caption{Comparison of classification performance between conventional CNN models and ProtoCAM under a 3-way 5-shot scenario.}
\label{fig2:proto_cnn}
\end{figure}

A conventional CNN trained on large-scale labelled images will outperform few-shot models, but that is not the scenario we are evaluating. The objective is to demonstrate the clinical value of few-shot learning in settings where well-annotated data is scarce, a constraint that is not the exception in the medical domain, but the norm. To validate the efficacy of episodic metric learning against conventional deep learning paradigms, we conducted a controlled benchmark experiment across seven CNN architectures. To maintain validity and eliminate confounding variables, all baselines were subjected to strict structural boundaries: they shared identical ImageNet initializations, consumed the exact same train/validation splits generated by the \texttt{StratifiedGroupKFold} pipeline, utilized uniform preprocessing parameters, and crucially, were trained under exact class support constraints where they only had access to the support samples per class within each fold. As illustrated in Figure~\ref{fig2:proto_cnn}, the performance discrepancies under this 3-way 5-shot scenario after 40 epochs reveal a stark divide between the two learning philosophies.

The conventional CNN baselines uniformly collapsed or struggled to generalize under these extreme data constraints. The standard ResNet18 classifier yielded the highest baseline performance with a Macro F1-score of 0.531, followed closely by EfficientNetB1 at 0.517 and MobileNetV2 at 0.471. As expected, scaling the network depth severely degraded baseline performance; ResNet50 and ResNet101 stalled at Macro F1-scores of 0.419 and 0.411, respectively. Conversely, ProtoCAM utilizing the exact same ResNet18 backbone achieved a Macro F1-score of 0.910 under identical few-shot constraints, outperforming its direct supervised counterpart by an absolute margin of 37.9\%. These empirical results demonstrate that achieving clinical-grade accuracy on scarce ultrasound data may require a shift away from traditional supervised cross-entropy optimization and toward sample-efficient architectures.

\subsection{Visual Interpretation of ProtoCAM Predictions}
\begin{figure}[!htbp]
    \centering
    \captionsetup[subfigure]{justification=centering} 

    \begin{subfigure}[b]{0.35\textwidth}
        \centering
        \includegraphics[width=\textwidth]{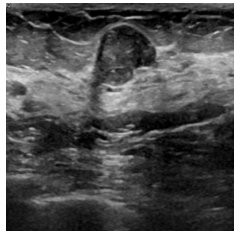}
        \caption{Benign, original}
    \end{subfigure}
    \begin{subfigure}[b]{0.35\textwidth}
        \centering
        \includegraphics[width=\textwidth]{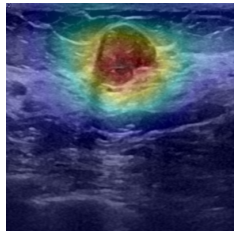}
        \caption{Benign, annotated}
    \end{subfigure}

    \begin{subfigure}[b]{0.35\textwidth}
        \centering
        \includegraphics[width=\textwidth]{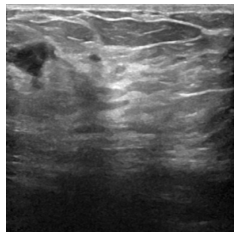}
        \caption{Malignant, original}
    \end{subfigure}
    \begin{subfigure}[b]{0.35\textwidth}
        \centering
        \includegraphics[width=\textwidth]{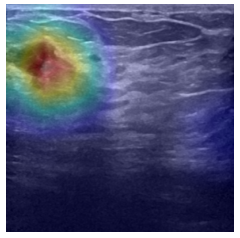}
        \caption{Malignant, annotated}
    \end{subfigure}

    \begin{subfigure}[b]{0.35\textwidth}
        \centering
        \includegraphics[width=\textwidth]{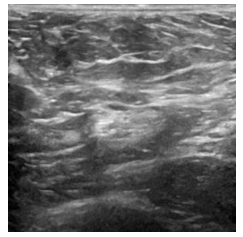}
        \caption{Normal, original}
    \end{subfigure}
    \begin{subfigure}[b]{0.35\textwidth}
        \centering
        \includegraphics[width=\textwidth]{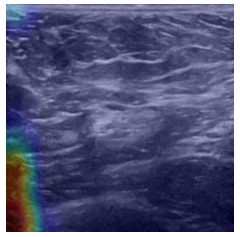}
        \caption{Normal, annotated}
    \end{subfigure}
    
    \caption{Examples of attention maps produced by ProtoCAM. Each row shows a different breast ultrasound image, with the original input image on the left and the corresponding Grad-CAM visualization on the right. The first row illustrates a benign case, the second row a malignant case, and the third row a normal case.}
    \label{fig3:gradcam}
\end{figure}

Figure~\ref{fig3:gradcam} presents representative Grad-CAM visualizations generated by the ProtoCAM model with a ResNet18 backbone trained under the 3-way 15-shot setting with a query size of 25 (the best-performing model, see Table~\ref{tab3:shots}). Each pair of images shows the original ultrasound image and the corresponding Grad-CAM heatmap highlighting the regions that most strongly influence the model’s prediction. In the benign case (a–b), the Grad-CAM map concentrates primarily around the lesion region, with strong activation localized within the mass boundary. The surrounding background tissue receives minimal attention, indicating that the model relies mainly on the relevant structural characteristics of the lesion for classification. For the malignant case (c–d), the activation map again highlights the lesion area. The model appears to focus on the irregular margins and internal texture patterns of the lesion, suggesting that these features play a key role in the classification decision.

In the normal case (e–f), the Grad-CAM visualization shows relatively diffuse and low-intensity activation across the image, without a strong localized focus. This may indicate that the model does not detect a suspicious region resembling a lesion. These visualizations demonstrate that ProtoCAM produces clinically meaningful explanations by focusing attention on anatomically relevant regions. Specifically, in lesion-containing images, the model concentrates on the tumour area. 

\section{Conclusion}\label{sec:conclusion}
This study presented ProtoCAM, an explainable few-shot learning framework for breast cancer classification in ultrasound imaging that integrates mask-guided feature encoding, prototypical metric learning, and Grad-CAM–based ~interpretability. The proposed approach was evaluated on the BUSI dataset using a rigorous Stratified Group K-Fold cross-validation protocol to prevent patient-level data leakage. Experimental results demonstrated that ProtoCAM significantly outperforms conventional CNN classifiers under severe data constraints, achieving a macro F1-score of 0.910 in a 3-way 5-shot setting. 

Among the evaluated backbone architectures, ResNet18 provided the best balance between representation capacity and generalization, achieving the highest macro F1-score of 91.65 percent. Sensitivity analysis of episodic parameters further showed that a 15-shot configuration with 25 query samples per class produced the best performance, indicating an optimal balance between prototype stability and training diversity. Additionally, Grad-CAM visualizations confirmed that the model focuses on clinically meaningful lesion regions rather than background artifacts, supporting the interpretability of the framework. The results demonstrate that the proposed ProtoCAM architecture provides an effective and interpretable solution for breast ultrasound analysis in data-scarce medical imaging environments, highlighting the potential of few-shot metric learning for reliable computer-aided diagnosis.

\section{Limitations and Future Work}\label{sec:limit}
Despite the promising results, several limitations should be acknowledged. First, the experiments were conducted on a single public dataset (BUSI), which may limit the generalizability of the proposed ProtoCAM framework to ultrasound data acquired from different institutions, imaging devices, or patient populations. Second, the approach relies on the availability of lesion masks for mask-guided feature encoding, which may not always be readily available in real clinical settings and typically require manual annotation by experts. Future work will therefore focus on evaluating the model across multiple multi-centre ultrasound datasets to assess its robustness and generalization capability. Additionally, incorporating automatic or weakly supervised lesion localization methods could reduce the dependence on manual annotations.

\bibliographystyle{plain}

\end{document}